\nonstopmode
\documentclass[a4paper]{article}

\usepackage{INTERSPEECH2015}

\usepackage{graphicx}
\usepackage{amssymb,amsmath,bm}
\usepackage{lipsum}
\usepackage{textcomp}

\ninept

\title{Deep Recurrent Neural Networks for Acoustic Modelling}

\makeatletter
\def\name#1{\gdef\@name{#1\\}}
\makeatother \name{{\em William Chan$^1$, Ian Lane$^{1,2}$}}

\address{Carnegie Mellon University \\
  $^1$Electrical and Computer Engineering, $^2$Language Technologies Institute \\
  {\small \tt williamchan@cmu.edu, lane@cmu.edu}
}

\begin{document}

  \maketitle
  \begin{abstract}
    We present a novel deep Recurrent Neural Network (RNN)
model for acoustic modelling in Automatic Speech Recognition
(ASR). We term our contribution as a TC-DNN-BLSTM-DNN
model, the model combines a Deep Neural Network (DNN)
with Time Convolution (TC), followed by a Bidirectional LongShort
Term Memory (BLSTM), and a final DNN. The first DNN
acts as a feature processor to our model, the BLSTM then generates
a context from the sequence acoustic signal, and the final
DNN takes the context and models the posterior probabilities of
the acoustic states. We achieve a 3.47 WER on the Wall Street
Journal (WSJ) eval92 task or more than 8\% relative improvement
over the baseline DNN models
  \end{abstract}
  
  \noindent{\bf Index Terms}: Deep Neural Networks, Recurrent Neural Networks, Long-Short Term Memory, Asynchronous Stochastic Gradient Descent, Automatic Speech Recognition

  \section{Introduction}
  Deep Neural Networks (DNNs) and Convolutional Neural Networks (CNNs) have yielded many state-of-the-art results in acoustic modelling for Automatic Speech Recognition (ASR) tasks \cite{jaitly-interspeech-2012, sainath-asru-2013}. DNNs and CNNs often accept some spectral feature (e.g., log-Mel filter banks) with a context window (e.g., +/- 10 frames) as inputs and trained via supervised backpropagation with softmax targets learning the Hidden Markov Model (HMM) acoustic states.
  
  DNNs do not make much prior assumptions about the input feature space, and consequently the model architecture is blind to temporal and frequency structural localities. CNNs are able to directly model local structural localities through the usage of convolutional filters. CNN filters connect to only a subset region of the feature space and are tied and shared across the entire input feature, giving the model translational invariance \cite{sainath-icassp-2013}. Additionally, pooling is often added, which yields rotational invariance \cite{sainath-asru-2013}. The inherent structure of CNNs yields a model much more robust to small shifts and permutations.
  
  Speech is fundamentally a sequence of time signals. CNNs (with time convolution) can capture some of this time locality through the convolution filters, however CNNs may not be able to directly capture longer temporal signal patterns. For example, temporal patterns may span 10 or more frames, however the convolution filter width may only be 5 frames wide. The CNN model must then rely on the higher level fully connected layers to model these long term dependencies. Additionally, one size may not fit all, the frame width of phones and temporal patterns are of varying lengths. Optimizing the convolution filter size is a expensive procedure and corpora dependent \cite{chan-icassp-2015}.
  
  Recently, Recurrent Neural Networks (RNNs) have been introduced demonstrating power modelling capabilities for sequences \cite{graves-icml-2012, graves-icassp-2013, graves-asru-2013, sak-interspeech-2014}. RNNs incorporate feedback cycles in the network architecture. RNNs include a temporal memory component (for example, in LSTMs the cell state \cite{hochreiter-neuralcomputation-1997}), which allows the model to store temporal contextual information directly in the model. This relieves us from explicitly defining the size of temporal contexts (e.g., the time convolution filter size in CNNs), and allows the model to learn this directly. In fact in \cite{sak-interspeech-2014}, the whole speech sequence can be accumulated in the temporal context.
  
  There exist many implementations of RNNs \cite{chung-nips-2014}. LSTM and Gated Recurrent Units (GRUs) \cite{chung-nips-2014} are particular implementations of RNNs that are easy to train and do not suffer from the vanishing or exploding gradient problems when performing Backpropagation Through Time (BPTT) \cite{hochreiter-2011}. LSTMs have the capability to remember sequences with long range temporal dependencies \cite{hochreiter-neuralcomputation-1997} and have been applied successfully to many applications include image captioning \cite{vinyals-arvix-2014}, end-to-end speech recognition \cite{graves-icml-2014} and machine translation \cite{sutskever-nips-2014}.

    LSTMs process sequential signals in one direction. One natural extension is bidirectional LSTMs (BLSTMs), which is composed of two LSTMs. The forward LSTM process the sequence as usual (e.g., reads the input sequence in the forward direction), the second processes the input sequence in backward order. The outputs of the two sequences can then be concatenated. BLSTMs have two distinct advantages over LSTMs, the first advantage being the forward and backward passes of the sequence yields differing temporal dependencies, the model can capture both sets of the signal dependencies. The second advantage is the higher level sequence layers (e.g., stacked BLSTMs) using the BLSTM outputs can access information from both input directions.
    
    LSTMs and GRUs (and their bidirectional variants) have recently been successfully applied to acoustic modelling and ASR \cite{graves-icml-2012, graves-asru-2013, sak-interspeech-2014}. In \cite{graves-icml-2012} TIMIT phone sequences were trained end-to-end from unsegmented sequence data using a LSTM transducer. LSTMs can be combined with Connectionist Temporal Classification (CTC) and implicitly perform sequence training over the speech signal on TIMIT \cite{graves-asru-2013}. \cite{chorowski-nips-2014} used GRUs and generated an explicit alignment model between the TIMIT speech sequence data to the phone sequence. In \cite{sak-interspeech-2014} a commercial speech system is trained using a LSTM acoustic model, here the the entire speech sequence is used as the context for classifying context dependent phones. \cite{sak-interspeech-2014b} extend from \cite{sak-interspeech-2014} and applied sequence training on top of LSTMs. Our contribution in this paper is a novel deep RNN acoustic model which is easy to train and archives an 8\% relative improvement over DNNs for the Wall Street Journal (WSJ) corpus.

  \section{Model}

\begin{figure*}[t]
  \includegraphics[width=\linewidth]{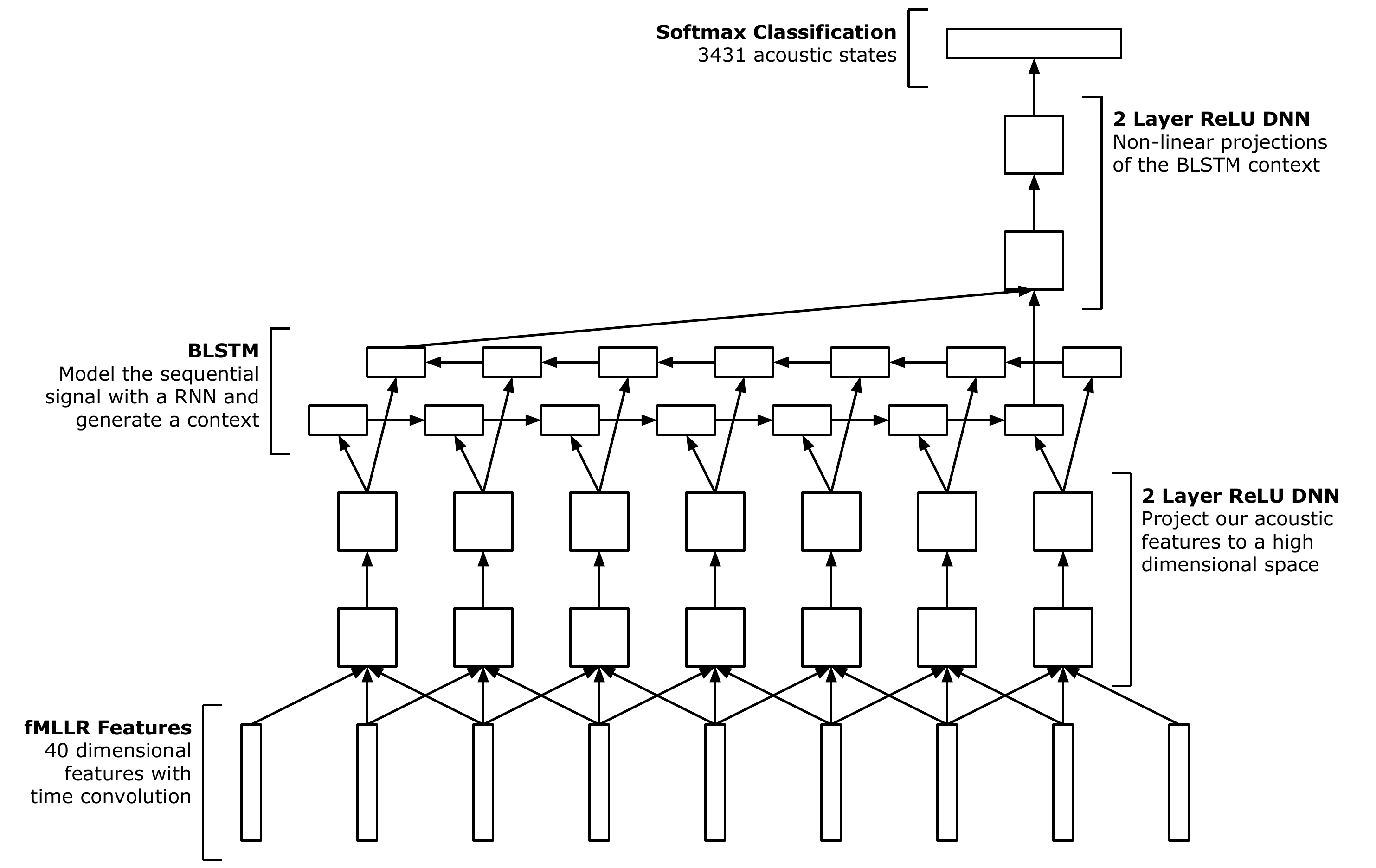}
\caption{TC-DNN-BLSTM-DNN Architecture. The model contains 3 parts, a signal processing DNN which takes in the original fMLLR acoustic features and projects them to a high dimensional space, a BLSTM which models the sequential signal and produces a context, and a final DNN which takes the context generated by the BLSTM and estimates the likelihoods across acoustic states.}
  \label{fig:architecture}
\end{figure*}
    Our model architecture can be summarized as a TC-DNN-BLSTM-DNN acoustic model. Our model deals with fixed length sequences (as opposed to variable length whole sequences \cite{sak-interspeech-2014}) of a context window. The advantage of our model is we can easily use BLSTMs online (e.g., we don't need to wait to see the end of the sequence to generate the backward direction pass of the LSTM). The disadvantage is however the amount of temporal information stored in the model is limited to the context width (e.g., similar to DNNs and CNNs). However, in offline decoding, we can also compute all the acoustic states in parallel (e.g., one big minibatch) versus the $O(T)$ iterations needed by \cite{sak-interspeech-2014} due to the iterative dependency of the LSTM memory.
    
    The model begins with a fixed window context of acoustic features (e.g., fMLLR) similar to a standard DNN or CNN acoustic model \cite{hinton-ieeespm-2012, sainath-icassp-2013}. Within the context window, an overlapping time window of features, or Time Convolution (TC) of features is fed in at each timestep. A similar approach was used by \cite{awni-arxiv-2014}, however they used a stride of 2 for the sake of reducing computational cost, however, our motivation is time convolution rather than performance and we use a stride of 1. The model processes these features with independent columns of DNNs over the context window timesteps. We refer this as the TC-DNN component of the model. The objective of the TC-DNN component is to project the original acoustic feature into a high dimensional feature space which can then be easily modelled or consumed by the LSTM. \cite{pascanu-iclr-2014} refers to this as a Deep Input-to-Hidden Function.

The transformed high dimensional acoustic signal is then fed into a BLSTM. The BLSTM models the time sequential component of the signal. Our LSTM implementation is similar to \cite{vinyals-arvix-2014} and described in the equations below:
\begin{align}
  i_t &= \phi(W_{xi} x_t + W_{hi} h_{t - 1}) \\
  f_t &= \phi(W_{xf} x_t + W_{hf} h_{t - 1}) \\
  c_t &= f_t \odot cs_{t - 1} + i_t \odot \tanh(W_{xc} x_t + W_{hc} h_{t - 1}) \\
  o_t &= \phi(W_{xo} x_t + W_{ho} h_{t - 1}) \\
  h_t &= o_t \odot \tanh(c_t)
\end{align}

We do not use bias, nor peephole connections; on initial experimentation, we observed negligible difference, hence we omitted them in this work. Additionally, we did not apply any gradient clipping or gradient projection, we did however apply a cell activation clipping of 3 to prevent saturation in the sigmoid non-linearities. We found the cell activation clipping to help remove convergence problems and exploding gradients. We also do not use a recurrent projection layer \cite{sak-interspeech-2014}. We found our LSTM implementation to train very easily without exploding gradients, even with high learning rates.

The BLSTM scans our input acoustic window of time width $T$ emitted by the first DNN and outputs two fixed value vector (one for each direction), which is then concatenated:
\begin{align}
c = \begin{bmatrix} h_T^f \\ h_1^b \end{bmatrix}
\end{align}
We refer $c$ as the context of the acoustic signal generated by the BLSTM. Context $c$ compresses the relevant acoustic information needed to classify the phones from the feature context (e.g., the window of fMLLR features).

The context is further manipulated and projected by a second DNN. The second DNN adds additional non-linear transformations before being finally fed to the softmax layer to model the context dependent state posteriors. \cite{pascanu-iclr-2014} refers this as the Deep Hidden-to-Output Function. The model is trained supervised with backpropagation minimizing the cross entropy loss. Figure \ref{fig:architecture} gives a visualization of our entire model.

  \section{Optimization}
    We found our LSTM models to be very easy to train and converge. We initialize our LSTM layers with a uniform distribution $\mathcal{U}(-0.01, 0.01)$, and our DNN layers with a Gaussian distribution $\mathcal{N}(0, 0.001)$. We clip our LSTM cell activations to 3, we did not need to apply any gradient clipping or gradient projection.
    
    We train our model with Stochastic Gradient Descent (SGD) using a minibatch size of 128, we found using larger minibatches (e.g., 256) to give slightly worse WERs. We used a simple geometric decay schedule, we start with a learning rate of 0.1 and multiply it by a factor of 0.5 every epoch. We have a learning rate floor of $0.00001$ (e.g., the learning rate does not decay beyond this value). We experimented with both classical and Nesterov momentum, however we found momentum to harm the final WER convergence slightly, hence we use no momentum. We apply the same optimization hyperparameters for all our experiments, it is possible using a slightly different decay schedule will yield better results. Our best model took 17 epochs to converge or around 51 hours in wall clock time with a NVIDIA Tesla K20 GPU.

  \section{Experiments and Results}
    We experiment with the WSJ dataset. We use si284 with approximately 81 hours of speech as the training set, dev93 as our development set and eval92 as our test set. We observe the WER of our development set after every epoch, we stop training once the development set no longer improves. We report the converged dev93 and the corresponding eval92 WERs. We use the same fMLLR features generated from the Kaldi s5 recipe \cite{povey-asru-2011}, and our decoding setup is exactly the same as the s5 recipe (e.g., large dictionary and trigram pruned language model). We use the tri4b GMM alignments as our training targets and there are a total of 3431 acoustic states. The GMM tri4b baseline achieved a dev and test WER of 9.39 and 5.39 respectively.
  
  \subsection{DNN}
  
    Two baseline DNN systems are presented, the first is the Kaldi s5 WSJ recipe with sigmoid DNN model which pretrains with a Deep Belief Network \cite{hinton-neuralcomputation-2006}, it achieved a WER of 3.81.
    
    We also built a ReLU DNN which requires no pretraining. The ReLU DNN consisted of 4 layers of 2048 ReLU neurons followed by softmax and trained with geometrically decayed SGD. We also experimented with deeper and wider networks, however we found this 5 layer architecture to be the best. Our ReLU DNN is much easier to train (e.g., no expensive pretraining) and achieves a WER of 3.79 matching the WER of the pretrained Sigmoid DNN. The ReLU DNN results suggest that pretraining may not be necessary given sufficient supervised data and is competitive for the acoustic modelling task. Table \ref{tab:dnn} summarizes the WERs for our DNN baseline systems.
    
    \begin{table}[h]
\caption{\label{tab:dnn} {\it WERs for Wall Street Journal. The ReLU DNN requires no pretraining and matches the WER of the Kaldi s5 recipe which uses DBN pretraining.}}
\vspace{2mm}
\centerline{
\begin{tabular}{|c|c|c|}
\hline
\bfseries Model & dev93 WER & eval92 WER \\
\hline
\hline
GMM Kaldi tri4b & 9.39 & 5.39 \\
DNN Kaldi s5  & 6.68 & 3.81 \\
DNN ReLU      & 6.84 & 3.79 \\
\hline
\end{tabular}
}
\end{table}

    \subsection{Deep BLSTM}

   We experimented with single layer and two layer deep BLSTM models. The cell size reported is per direction (e.g., total cells are doubled). The BLSTM models take longer to train and underperform compared to the ReLU DNN model. The large BLSTM models tend to outperform the smaller ones, suggesting overfitting is not an issue. However, there is limited incremental gain in WER performance with additional cells. Our best single layer BLSTM with 1024 bidirectional cells achieved only 4.06 WER compared to 3.79 from our ReLU DNN model.
   
   Deep BLSTM models \cite{graves-asru-2013} may give additional model performance, since the upper layers can access information from the shallow layers in both directions and additional layers of non-linearities are available. Our deep BLSTM models contain two layers, the cell size reported is per direction per layer (e.g., total cells are quadrupled). Our deep BLSTM experiments give mixed results. For the same number of cells per layer, the deep model performs slightly better. However, if we fixed the number of parameters, the single layer BLSTM model performs slightly better, the single layer of 1024 bidirectional cells achieved a WER of $4.16$ while the deep two layer BLSTM model with 512 bidirectional cells per layer achieved a WER $4.25$.  Table \ref{tab:blstm} summarizes our BLSTM experiment WERs.
   
   \begin{table}[t]
        \caption{\label{tab:blstm} {\it BLSTM WERs for Wall Street Journal. Larger recurrent models tend to perform better without overfitting. The deep BLSTM models do not yield any substantial gains over their single layer counterparts.}}
        \vspace{2mm}
        \centerline{
        \begin{tabular}{|c|c|c|c|}
        \hline
        \bfseries Cell Size & \bfseries Layers & \bfseries dev93 WER & \bfseries eval92 WER \\
        \hline
        \hline
        128 & 1 & 8.19 & 5.19 \\
        256 & 1 & 7.94 & 4.66 \\
        512 & 1 & 7.43 & 4.36 \\
        768 & 1 & 7.36 & 4.16 \\
        1024 & 1 & 7.23 & 4.06 \\
        256 & 2 & 7.54 & 4.36 \\
        512 & 2 & 7.40 & 4.25 \\
        \hline
        \end{tabular}
        }
    \end{table}
   
   \subsection{TC-DNN-BLSTM-DNN}
   
   We experimented next with a DNN-BLSTM model. Our DNN-BLSTM model does not have time convolution at its input, and lacks the second DNN non-linearities for context projection. The two layer 2048 neuron ReLU DNN in front of the BLSTM acts as a signal processor, projecting the original acoustic signal (e.g., each fMLLR vector) into a new high dimensional space which can be more easily digested by the LSTM. The BLSTM module uses 128 bidirectional cells. Compared to the 128 bidirectional cell BLSTM model, the model improves from 5.19 WER to 3.92 WER or 24\% relatively. The results of this experiment suggest the fMLLR features may not be the best features for BLSTM models (to consume directly at least); but rather learnt features (through the DNN feature processor) can yield better features for the BLSTM model to consume.
   
   The next experiment we ran was a BLSTM-DNN model. Here, the BLSTM accepts the original acoustic feature without modification and emits a context. The context is passed through to a two layer 2048 neuron ReLU DNN which provides additional layers of non-linear projections before classification by the softmax layer. Once again, the BLSTM module uses only 128 bidirectional cells. The model improves from 5.19 WER to 3.84 WER or 26\% relatively when compared to the original 128 bidirectional cell BLSTM model which does not have the context non-linearities. The result of this experiment suggest the LSTM context should not be used directly for softmax phone classification, but rather additional layers of non-linearities are needed to achieve the best performance.
   
   We then experimented with a DNN-BLSTM-DNN model (without time convolution). Each DNN has two layers of 2048 ReLU neurons, and the BLSTM layer had 128 cells per direction. We combine both the benefits of a learnt signal processing DNN and the context projection. Compared to a 128 bidirectional cell BLSTM model, our WER drops from 5.19 to 3.76 or 28\% relatively. Compared to a 1024 bidirectional cell BLSTM model, we essentially redistributed our parameters from a wide shallow network to a deeper network. We achieve a 11\% relative improvement compared to a single layer 1024 bidirectional cell BLSTM, suggesting the deeper models are much more expressive and powerful.

    Finally, our TC-DNN-BLSTM-DNN model combines the DNN-BLSTM-DNN with input time convolution. Our model further improves from 3.76 WER without time convolution to 3.47 WER with time convolution. Compared to the DNN models, we achieve 0.32 absolute WER reduction or 8\% relatively. To the best of our knowledge, this is the best WSJ eval92 performance without sequence training \cite{vesely-interspeech-2013}. We hypothesize the time convolution gives a richer signal representation to the DNN signal processor and consequently the BLSTM model to consume. The time convolution also relieves the LSTM computation power to learning long term dependencies, rather than short term dependencies. Table \ref{tab:dnnblstmdnn} summarizes the experiments for this section.

\begin{table}[t]
\caption{\label{tab:dnnblstmdnn} {\it Ablation effects of our TC-DNN-BLSTM-DNN model. The DNNs and Time Convolution are used for signal and context projections. We show that all components are critical to obtain the best performing model.}}
\vspace{2mm}
\centerline{
\begin{tabular}{|c|c|c|}
\hline
\bfseries Model & dev93 WER & eval92 WER \\
\hline
\hline
DNN-BLSTM        & 7.40 & 3.92 \\
BLSTM-DNN        & 6.90 & 3.84 \\
DNN-BLSTM-DNN    & 7.19 & 3.76 \\
TC-DNN-BLSTM-DNN & 6.58 & 3.47 \\
\hline
\end{tabular}
}
\end{table}

\subsection{Distributed Optimization}
    All results presented in the previous sections of this paper were trained with a single GPU with SGD. To reduce the time required to train an individual model we also experimented with distributed Asynchronous Stochastic Gradient Descent (ASGD) across multiple GPUs. Our implementation is similar to \cite{chan-interspeech-2014}, we have 4 GPUs (NVIDIA Tesla K20) in our system, 1 GPU is dedicated as a parameter server and we have 3 GPU compute shards (e.g., the independent SGD learners). We do not apply any stale gradient decay \cite{chan-interspeech-2014} or warm starting \cite{dean-nips-2012}. We use the exact same learning rate schedule, minibatch size and hyperparameters as our TC-DNN-BLSTM-DNN SGD baseline. \cite{sak-interspeech-2014} applied distributed ASGD optimization, however they applied it on a cluster of CPUs rather than GPUs. Additionally, \cite{sak-interspeech-2014} did not compare if there was a WER differential between SGD versus ASGD.
 
\begin{figure}[t]
      \includegraphics[width=\linewidth]{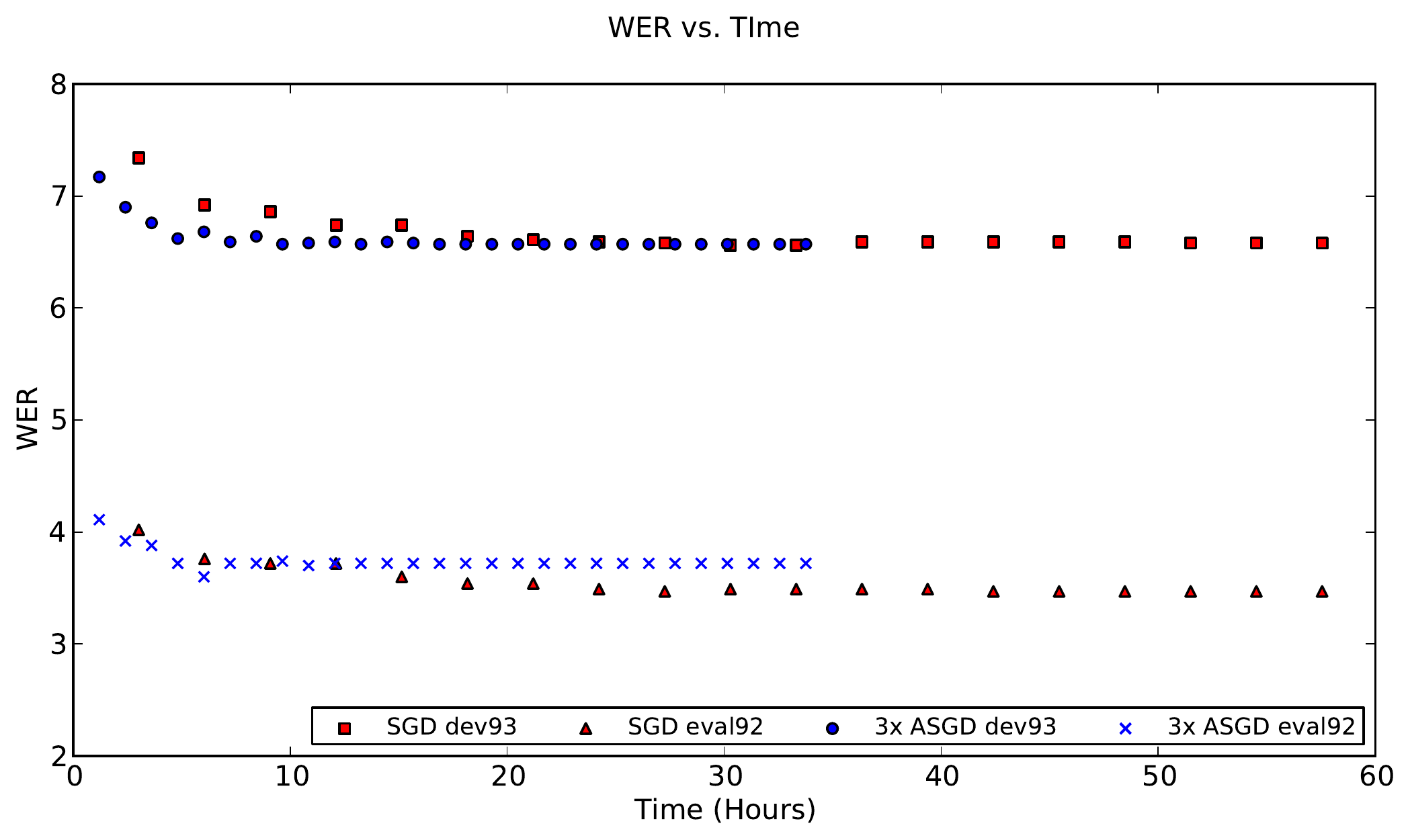}
      \caption{SGD vs x3 ASGD WER convergence comparison, each point represents one epoch of the respective optimizer.}
      \label{fig:distributed}
    \end{figure}

    Our baseline TC-DNN-BLSTM-DNN SGD system took 17 epochs or 51 wall clock hours to converge to a dev and test WER of 6.58 and 3.47. Our distributed implementation converges in 14 epochs and 16.8 wall clock hours, achieves a dev and test WER of 6.57 and 3.72. The distributed optimization is able to match the dev WER, however the test WER is significantly worse. It is unclear whether this WER differential is due to the asynchronicity characteristic of the optimizer or due to the small datasets, we suspect with larger datasets the gap between the ASGD and SGD will shrink. The conclusion we draw is that ASGD can converge much quicker and faster, however there may be a impact to final WER performance. Table \ref{tab:distributed} and Figure \ref{fig:distributed} summarizes our results.
    
    \begin{table}[t]
        \caption{\label{tab:distributed} {\it Effects of distributed optimization for our TC-DNN-BLSTM-DNN model. The ASGD experiments uses 3 independent SGD shards.}}
        \vspace{2mm}
        \centerline{
        \begin{tabular}{|c|c|c|c|c|}
        \hline
        \bfseries Model & Epochs & Time (hrs) & dev93 WER & eval92 WER \\
        \hline
        \hline
        SGD & 17 & 51.5 & 6.58 & 3.47 \\
        ASGD & 14 & 16.8 & 6.57 & 3.72 \\
        \hline
        \end{tabular}
        }
    \end{table}   
    
  \section{Conclusions}
    In this paper, we presented a novel TC-DNN-BLSTM-DNN acoustic model architecture. On the WSJ eval92 task, we report a 3.47 WER or more than 8\% relative improvement over the DNN baseline of 3.79 WER. Our model is easy to optimize and implement, and does not suffer from exploding gradients even with high learning rates. We also found that pretraining may not be necessary for DNNs, the DBN pretrained DNN achieved a 3.81 WER compared to our ReLU DNN without pretraining of 3.79 WER. We also experimented with ASGD with our TC-DNN-BLSTM-DNN model, we were able to match the SGD dev WER, however the WER on the evaluation set was significantly lower at 3.72. In future work, we seek to apply sequence training on top of our acoustic model to further improve the model accuracy.

  
  \eightpt
  \bibliographystyle{IEEEtran}
  \bibliography{paper}
\end{document}